\documentclass[11pt,a4paper]{article}
\usepackage{times,latexsym}
\usepackage{url}
\usepackage[T1]{fontenc}

\usepackage[acceptedWithA]{tacl2021v1}

\usepackage{xspace,mfirstuc,tabulary}

\usepackage{graphicx}
\usepackage{CJKutf8}
\usepackage{subfig}
\usepackage{booktabs, multirow} % for borders and merged ranges
\newif\iftaclinstructions
\taclinstructionsfalse % AUTHORS: do NOT set this to true
\iftaclinstructions
\renewcommand{\confidential}{}
\renewcommand{\anonsubtext}{(No author info supplied here, for consistency with
TACL-submission anonymization requirements)}
\newcommand{\instr}
\fi

\iftaclpubformat % this "if" is set by the choice of options

\else

\fi

\title{AutoCBT: An Autonomous Multi-agent Framework for Cognitive Behavioral Therapy in Psychological Counseling}

\author{
\bf
Ancheng Xu$^{1,2}$\footnotemark[1] \quad Di Yang$^{1,3}$\footnotemark[1] \quad Renhao Li$^{1,4}$\footnotemark[1] \quad Jingwei Zhu$^{1,3}$\footnotemark[1] \quad Minghuan Tan$^{1}$\footnotemark[2] \quad Min Yang$^{1}$\footnotemark[2]\\
\bf
Wanxin Qiu$^{5}$ \quad Mingchen Ma$^{1}$ \quad Haihong Wu$^{1,3}$ \quad Bingyu Li$^{5}$ \quad Feng Sha$^{1}$\\
\bf
Chengming Li$^{6}$ \quad Xiping Hu$^{6}$ \quad Qiang Qu$^{1}$ \quad Derek F.Wong$^{4}$ \quad Ruifeng Xu$^{7}$\\
$^1$Shenzhen Key Laboratory for High Performance Data Mining, \\Shenzhen Institutes of Advanced Technology, Chinese Academy of Sciences \\
$^2$University of Chinese Academy of Sciences, $^3$University of Science and Technology of China\\
$^4$University of Macau, $^5$Shenzhen University, $^6$Shenzhen MSU-BIT University\\
$^7$Harbin Institute of Technology, Shenzhen\\
\texttt{\{ac.xu,mh.tan,min.yang\}@siat.ac.cn}\\
\texttt{\{di-yang,jingweizhu\}@mail.ustc.edu.cn}\\
\texttt{li.renhao@connect.um.edu.mo, qiuwanxin2023@email.szu.edu.cn}
}

\date{}

\begin{document}
\maketitle

% 这行命令将脚注编号的显示格式更改为 符号形式，而不是默认的阿拉伯数字（1, 2, 3）。
\renewcommand{\thefootnote}{\fnsymbol{footnote}}
\footnotetext[1]{Equal contribution.}
\footnotetext[2]{Corresponding authors.}

% 在这里重置脚注计数器，并将脚注编号恢复为数字形式
\setcounter{footnote}{0}
\renewcommand{\thefootnote}{\arabic{footnote}}

\begin{CJK*}{UTF8}{gbsn}
\begin{abstract}
Traditional in-person psychological counseling remains primarily niche, often chosen by individuals with psychological issues, while online automated counseling offers a potential solution for those hesitant to seek help due to feelings of shame. Cognitive Behavioral Therapy (CBT) is an essential and widely used approach in psychological counseling. The advent of large language models (LLMs) and agent technology enables automatic CBT diagnosis and treatment. However, current LLM-based CBT systems use agents with a fixed structure, limiting their self-optimization capabilities, or providing hollow, unhelpful suggestions due to redundant response patterns. In this work, we utilize Quora-like\footnote{\url{https://www.quora.com}} and YiXinLi\footnote{\url{https://www.xinli001.com/qa}} single-round consultation models to build a general agent framework that generates high-quality responses for single-turn psychological consultation scenarios. We use a bilingual dataset to evaluate the quality of single-response consultations generated by each framework. Then, we incorporate dynamic routing and supervisory mechanisms inspired by real psychological counseling to construct a CBT-oriented autonomous multi-agent framework, demonstrating its general applicability. Experimental results indicate that AutoCBT can provide higher-quality automated psychological counseling services.
\end{abstract}

\begin{figure*}[!htb]
\centering
\includegraphics[width=\linewidth]{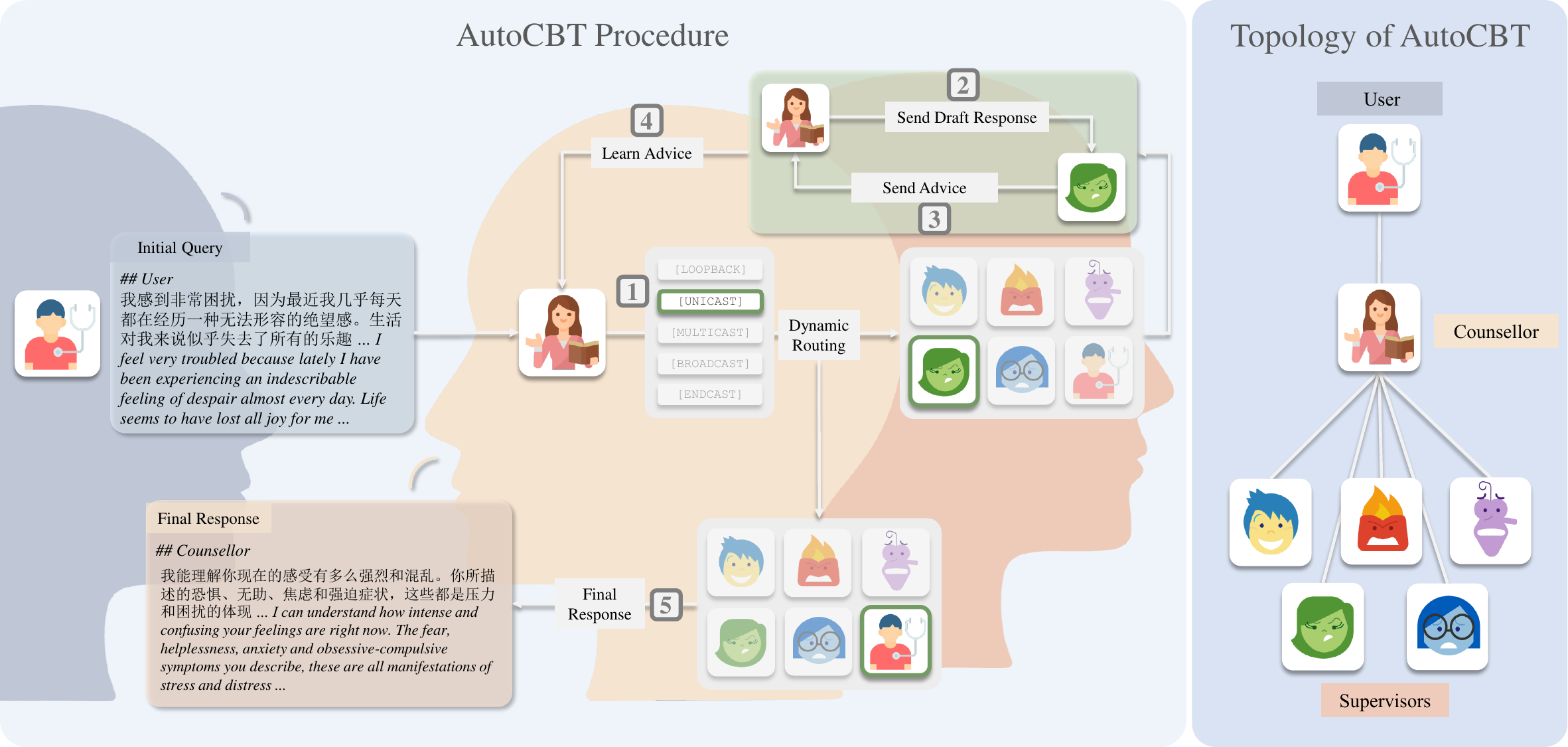}
\caption{Overview of the AutoCBT framework. Upon receiving a user's question, the counsellor first uses dynamic routing to determine whether to consult a supervisor or respond directly to the user. If the decision is to consult the supervisor, the counsellor drafts a response and sends it for advice. After receiving advice, the counsellor learns it and re-enters the dynamic routing process. If the counsellor responds directly to the user, the final response is immediately sent in step 5.}
\label{fig:framework}
\end{figure*}

\section{Introduction}
The field of psychological counseling has seen significant advancements with the integration of technology~\cite{althoff-etal-2016-large}, particularly through the use of large language models (LLMs)~\cite{demszky2023using,yang-etal-2023-towards}. 
These models have been increasingly employed to assist in the diagnosis and treatment of various psychological conditions, offering promising avenues for personalized and scalable mental health care. 
However, despite the progress made, there remains a need for more adaptive and context-aware therapeutic systems that can effectively mimic the complexity of human interaction and cognitive processes.

Cognitive Behavioral Therapy (CBT) is recognized as one of the most effective and widely adopted approaches in psychological counseling.
\citet{beck1979cognitive, beck1993cognitive} posits that individuals experiencing emotional or psychological problems often exhibit \textit{logical errors}, such as catastrophizing, labeling, or minimizing positive affirmations.
These \textit{logical errors}, also referred to as \textit{thinking errors}, can distort an individual's perception of their situation, experiences, or personality, hinder daily functioning, and reduce their quality of life~\cite{hofmann2011introduction}.
From a practical intervention perspective, identifying, and challenging cognitive distortions are critical steps in mitigating adverse emotions and behaviors, thereby fostering self-health management in individuals~\cite{hofmann2011introduction}.

In recent years, there have been attempts to leverage LLMs to enhance CBT delivery.
Existing methods like CBT-LLM~\cite{na-2024-cbt-llm} typically involve the use of prompt-based in-context learning to analyze consultants' conditions and generate therapeutic responses.
To our best knowledge, CoCoA~\cite{lee2024cocoacbtbasedconversationalcounseling} constructs memory mechanisms for a single agent for retrieval-augmented generation and applies CBT techniques for identifying cognitive distortions inherent in the user’s statements.
While these approaches have shown promise, they often suffer from limitations such as a lack of personalization, an inability to adapt to changing patient needs, and a limited understanding of the dynamic nature of therapeutic interactions.

This paper introduces AutoCBT, a general autonomous multi-agent framework that addresses shortcomings of existing CBT approaches based on LLMs. 

Structurally, AutoCBT incorporates a routing mechanism and memory mechanism to enhance the autonomous ability of each agent.
Each message generated by AutoCBT undergoes a structured process of reasoning and editing to meet specific requirements.
From the perspective of CBT, AutoCBT allows for more flexible topology adaptation as therapeutic techniques evolve, and we also demonstrate its efficacy in other purely prompt-based psychological counseling framework.

We compare our framework and its realizations against existing LLM-based approaches and analyze how each realization performs over a bilingual dataset.

Our paper contributes in the following aspects:
\begin{enumerate}
    \item We present AutoCBT, a CBT-oriented autonomous multi-agent framework that is both flexible and highly configurable. It delivers superior responses in single-turn conversational counseling scenarios.
    \item We demonstrate AutoCBT's capacity to enhance response quality within other pure prompt-based counseling framework.
    \item We discuss and address the challenges introduced by dynamic routing and supervisory mechanisms, and analyze the Llama model's over-protection issue.
\end{enumerate}

\begin{table*}[!htbp]
\centering
\footnotesize
\scalebox{0.98}{
\begin{tabular}{cp{6.5cm}p{6.5cm}c}
\toprule[1.5pt]
\multirow{3}{*}{\textbf{Lang.}}  &\multicolumn{2}{c}{\textbf{Dataset Examples}} &\multirow{3}{*}{\textbf{Count}}\\\cmidrule{2-3}
&  \multicolumn{1}{c}{\textbf{Question Description}} & \multicolumn{1}{c}{\textbf{Answers}} &\\\midrule
\multirow{9}{*}{EN}  &{Me and my sister in law are both pregnant right now. And I’ve been noticing the inconsistency of level of care about our baby from my fiancée side of the fam. This situation really has me depressed, and unsure what to do. for starters my sister in law and that side of the family has made it a competition between the babies, I don’t want it to be a competition. It always who can do what first......} &{Thank you for explaining this situation. How unfortunate that this share joyous event is turning into a competition.The experience of being left out or ignored as part of the situation is what needs to be addressed. First, I would have a talk with your fiancé about what is happening and why. Does the family have a bias against the pregnancy because you were not married first? Is your fiancé on the outs with his family......} &\multirow{9}{*}{100}\\\midrule
\multirow{13}{*}{ZH}  &\scriptsize{总是要考虑很多问题，我感觉我活在世界上就没有意义？我感觉我自己在交朋友的这条路上总是很不顺，初一初二的时候跟别人抢，我总是抢不过，不知道为什么，我总是把自己的地位放的很低，只要她一生气，我就卑微的求她原谅我，不管她说什么我都同意我总是感觉我拒绝一次，她就会离开我，一直到了上初三，爸妈突然让我转班，说是为了我的学业，我总是在课上无缘无故的哭，我真的很难受，到了高一，我认识了一个女孩，我们特别能合得来，我就看到了希望，我以为一切都会改变，但后来我发现，她很受人喜欢，班里的所有女生都很喜欢她，而我只能默默的看着，两个月以后，她总是前一天对我还很好，但是后一天又莫名其妙的对我冷暴力...} &\scriptsize{抱抱～看到发生在你身上的事就像往事重现。请允许我以姐姐的口吻与你讲下我的故事。我在刚去外地读大学的时候，认识了一位我很欣赏的女生。独立精干，双商高，性格开朗，很勇敢。她是一位好舍友，也是一位好闺蜜。入学初就约定好一起考研，从那就形影不离。在她面前，我可以表现很勇敢，很积极，很仗义，因为是我好朋友啊。上学一起，学习一起，吃饭一起，活动一起，实验一起，自习一起，睡觉也一起。反正就好的像黏黏胶一样。问想到我都上大学的人了还会这么粘人。后来想想可能是我一个人在外地吧，聊得来就会很上瘾哈哈。那段时光过得很愉快，回忆起来暖暖的。我一直以为会一直这样下去，连交男朋友都是外地异地恋，这样不会耽误我和闺蜜的时间...}  &\multirow{13}{*}{100}\\
\bottomrule[1.5pt]
\end{tabular}
}
\caption{The overall structure and a Q\&A example from the bilingual dataset.}
\label{tab:qa_examples}
\end{table*}

\section{Related Work}
CBT is a widely recognized treatment for mental health conditions like anxiety, depression, and addiction.
A key component of CBT is teaching users to recognize and correct their thinking traps, helping them discard negative thought processes.
Cognitive distortions~\cite{beck1963thinking} are biased or irrational thinking patterns that lead individuals to misinterpret reality, resulting in negative emotions and behaviors~\cite{beck2020cognitive}.
Such distortions create a feedback loop of unhealthy thinking, often becoming automatic and emotionally triggered~\cite{beck1979cognitive}.
Therefore, it's essential to help users identify these patterns and adopt appropriate strategies.

Recently, there's been growing interest in applying artificial intelligence (AI) to study cognitive distortions.

To assist AI in recognizing distortions, annotated datasets and ontologies based on CBT principles have been created~\cite{rojas-barahona-etal-2018-deep, wang-etal-2023-c2d2}.

Computer-based CBT systems have been developed to make therapeutic care more accessible.
Examples include an affectively-aware virtual therapist for depression counseling~\cite{ring2016affectively} and the Woebot chatbot delivering CBT via decision trees~\cite{fitzpatrick2017delivering}.
However, earlier systems often relied on predefined responses and lacked natural conversational abilities.

With the advent of LLMs, there's increasing interest in leveraging advanced AI to enhance CBT delivery.
Frameworks like CBT-LLM~\cite{na-2024-cbt-llm} use prompt-based learning to generate therapeutic responses.
Systems like CoCoA~\cite{lee2024cocoacbtbasedconversationalcounseling} incorporate memory mechanisms and apply CBT techniques to identify cognitive distortions in user statements.
Other studies assess the conversational behavior of LLM therapists~\cite{chiu2024computationalframeworkbehavioralassessment}.
Thus, we aim to evaluate the readiness of conversational agents based on CBT in therapeutic settings and assess their behavior during therapy sessions.

\section{Methodology}

\subsection{AutoCBT}

AutoCBT is a general framework that acts as a proxy for different multi-agent systems in the backend.
The structure of the framework can be represented as $(a_0, S, \mathcal{T}, \Sigma)$ where:
\begin{itemize}
    \item $a_0$ is the Counsellor Agent acting as the interface for the multi-agent system.
    \item $S = \{a_{i}|i\in[1,N]\}$ is the set of supervisor agents that the Counsellor Agent can seeking information from.
    \item $\mathcal{T}$ is the topology of communicable agents.
    \item $\Sigma$ is the set of allowed routing strategies among agents.
\end{itemize}

\begin{table*}[t]\centering
\small
\scalebox{0.9}{
\begin{tabular}{lp{4.4cm}p{8.5cm}c}
\toprule[1.5pt]
\textbf{Perspective} &\textbf{Description} &\textbf{Criterion} &\textbf{Score} \\\midrule
\multirow{5}{*}{Empathy} &\multirow{5}{4.5cm}{Demonstrates understanding and sympathy towards the user's emotions or issues, and creates a sense of safety.} & 1.1 Did the counsellor correctly understand the user's intent? &\multirow{5}{*}{7} \\ 
~ & ~ & 1.2 Did the counsellor show respect, understanding, and sympathy for the user's anxiety and pain? & ~ \\
~ & ~ & 1.3 Did the counsellor create a safe environment for the user to express their feelings? & ~ \\ \midrule
\multirow{4}{*}{Identification} &\multirow{4}{4.5cm}{Identify potential cognitive distortions of the user through the description of the problem in the dialogue.} & 2.1 Did the counsellor identify the user's distorted beliefs? &\multirow{4}{*}{7} \\ 
~ & ~ & 2.2 Did the counsellor delve into the user's distorted beliefs? & ~ \\
~ & ~ & 2.3 Did the counsellor assist the user in recognizing and challenging these distorted beliefs? & ~ \\ \midrule
\multirow{6}{*}{Reflection} &\multirow{6}{4.5cm}{Ask open-ended questions to encourage the user to reconsider or reflect on their initial thoughts or beliefs.} & 3.1 Did the counsellor ask questions related to the user's initial thoughts? &\multirow{6}{*}{7} \\ 
~ & ~ & 3.2 Did the counsellor pose questions that facilitated deeper thinking? & ~ \\
~ & ~ & 3.3 Did the counsellor ask questions reflecting the user's distorted beliefs? & ~ \\ \midrule
\multirow{6}{*}{Strategy} &\multirow{6}{4.5cm}{Provide practical strategies or insights to help the user address their current situation.} & 4.1 Were the strategies or insights provided by the counsellor practical? &\multirow{6}{*}{7} \\ 
~ & ~ & 4.2 Could the strategies or insights solve the user's current problems? & ~ \\
~ & ~ & 4.3 Were the strategies based on professional psychological methods? & ~ \\ \midrule
\multirow{5}{*}{Encouragement} &\multirow{5}{4.5cm}{Encourage the user to use the strategies.} & 5.1 Did the counsellor encourage the user to take action? &\multirow{5}{*}{7} \\ 
~ & ~ & 5.2 Did the counsellor address potential failures the user might encounter while implementing the strategies? & ~ \\
~ & ~ & 5.3 Did the counsellor provide comfort and encouragement regarding setbacks and challenges? & ~ \\ \midrule
\multirow{5}{*}{Relevance} &\multirow{5}{4.5cm}{Evaluate the relevance of the dialogue content.} & 6.1 Was the counsellor's response highly relevant to the user's question? &\multirow{5}{*}{7} \\ 
~ & ~ & 6.2 Did the counsellor's response flow naturally? & ~ \\
~ & ~ & 6.3 Did the counsellor's answer cover the main issues or concerns raised by the user? & ~ \\ 
\bottomrule[1.5pt]
\end{tabular}
}
\caption{Six automatic evaluation metrics and corresponding score criterion based on the CBT core principles.}
\label{tab:metrics}
\end{table*}

\paragraph{Counsellor Agent} This agent is an interface for the AutoCBT multi-agent system which acts as the interface between the users with psychological confusion (either simulated users or real users from the web) and candidate supervisors.
The agent runs over an LLM and is configurable~(such as role description, routing prompt, message prompt, and so on) to automatically make decisions.

Upon receiving the original user question, it will forward the message based on professional judgment and seek extra information from different supervisor agents, until the counsellor agent has sufficient confidence to answer the user's question.

\paragraph{Supervisor Agents} Similar to the counsellor agent, these agents are also LLM-based and configurable.
However, the number of them and the way how they are connected can be modified according to the CBT approach adopted.

\paragraph{Memory Mechanisms} Each agent is accompanied by a short term memory to store most recent messages and a long-term memory to store summaries of messages with a sliding window.
\paragraph{Topology of agents} In AutoCBT, a topology is the graph~(either static or dynamic) of communicable agent pairs.
Messages can be transported over the topology but may endue subsequent modifications at each agent.
\paragraph{Routing Strategies} The routing strategies are defined for the commutable agents in the topology.
Generally speaking, there will be five different types of strategies which are defined as following:
\begin{enumerate}
    \item \texttt{[LOOPBACK]} Loop back, continue with the statement.
    \item \texttt{[UNICAST]} Unicast, send to a communicable agent.
    \item \texttt{[MULTICAST]} Multicast, send to several communicable agents.
    \item \texttt{[BROADCAST]} Broadcast, send to all communicable agents.
    \item \texttt{[ENDCAST]} Terminated casting, end communication with the specified agent.
\end{enumerate}

Such strategies enable agents with more autonomy to decide whether to seek information from other agents and in what order.

When the counsellor agent receives a message from real users or mocked users, it selects a communication target using allowed routing strategies and generates a response based on the configured prompts and conversation history.

Our framework is illustrated in Figure~\ref{fig:framework}.

\subsection{Decomposition of CBT Core Principles}

CBT-LLM~\cite{na-2024-cbt-llm} breaks down the CBT core principles into five standards: Validation and Empathy, Identify Key Thought or Belief, Pose Challenge or Reflection, Provide Strategy or Insight, Encouragement and Foresight.

In our topology, \textbf{the five standards will be projected onto the five supervisor agents, with each accounting for one standard}.
During inference, when the counsellor agent receives a message from the user, it makes a decision based on the message and its memory whether or not to seek information from one specific or multiple supervisors.
The process proceeds until the counsellor decides to reply to the user.

\subsection{Bilingual Dataset}
To validate the effectiveness of our proposed framework, we compile a bilingual dataset from existing counseling-oriented datasets PsyQA~\cite{sun-etal-2021-psyqa} and TherapistQA\footnote{\url{https://www.kaggle.com/datasets/arnmaud/therapist-qa}}, each focusing on Chinese and English respectively.

We use question descriptions from these datasets as the background context for the user.
To construct the bilingual dataset, we began by merging the questions and descriptions from PsyQA into unified instructions, simulating queries from real or hypothetical users directed toward the counsellor. 
We then employed a novel approach using large open-source language models, such as Qwen-72B~\cite{qwen} and LLaMA-70B~\cite{touvron2023llama}, to identify 10 cognitive distortions discussed in previous section within these concatenated instructions.

After classifying the instructions into their respective categories, we randomly selected 10 items from each class to create a dataset comprising 100 items, along with their corresponding answers, forming the Chinese and English splits of the bilingual dataset. 
Specific examples of the bilingual dataset can be found in Table~\ref{tab:qa_examples}.

\section{Experiments}
\subsection{Setting}
In online psychological counseling, LLMs must handle subtle emotional shifts and follow instructions effectively. We use \textbf{Qwen-2.5-72B} for Chinese and English sections and \textbf{Llama-3.1-70B} for English, with a temperature of 0.98 and default settings.

Baseline methods include \textbf{Generation}, where LLMs respond directly to bilingual dataset questions, and \textbf{PromptCBT}, which adds CBT principles to prompts before response generation.

AutoCBT operates in two stages: In the \textbf{Draft Response Process}, the counsellor generates a response and decides if supervisor assistance is needed. If required, the \textbf{Final Response Process} involves consulting a supervisor skilled in CBT principles to refine the response before sending it to the user.

\paragraph{Automatic Evaluation}
Based on Table~\ref{tab:metrics}, we use GPT-4o-mini automatic scoring to obtain our evaluation results. 
The form of the automatic scoring prompts for the six metrics remains similar, such as the scoring prompts for the reflective level of the consultant's answers.
Due to LLM response variability, each response is rated \textbf{three times} by GPT-4o-mini, and the scores are averaged in automatic scoring evaluation.

\begin{table*}[!htp]\centering
\centering
% \small
\scalebox{0.82}{
\begin{tabular}{cccccccc|c}
\toprule[1.5pt]
\multirow{2}{*}{\textbf{Lang.}} &\multirow{2}{*}{\textbf{Method}} &\multirow{2}{*}{\textbf{Empathy}} &\multicolumn{2}{c}{\textbf{Cognitive Distortions}} &\multirow{2}{*}{\textbf{Strategy}} &\multirow{2}{*}{\textbf{Encouragement}} &\multirow{2}{*}{\textbf{Relevance}}  &\multirow{2}{*}{\textbf{Total Score}}\\\cmidrule{4-5}
&&&\textbf{Identification}&\textbf{Reflection} &&&&\\\midrule

\multirow{3}{*}{ZH} &Generation  &5.493 / 7 &4.630 / 7 &4.280 / 7 &6.153 / 7 &5.200 / 7 &6.543 / 7 &32.300 \\
&PromptCBT  &6.000 / 7 &5.610 / 7 &5.623 / 7 &6.237 / 7 &6.130 / 7 &\textbf{6.860 / 7} &36.460  \\
&AutoCBT  &\textbf{6.247 / 7} &\textbf{5.760 / 7} &\textbf{5.787 / 7} &\textbf{6.363 / 7} &\textbf{6.447 / 7} &6.857 / 7 &\textbf{37.460}  \\\midrule

\multirow{3}{*}{EN} &Generation  &5.907 / 7 &4.903 / 7 &4.740 / 7 &6.093 / 7 &5.383 / 7 &6.637 / 7 &33.663 \\
&PromptCBT &6.390 / 7 &5.687 / 7 &5.797 / 7 &6.233 / 7 &6.377 / 7 &6.887 / 7 &37.370 \\
&AutoCBT   &\textbf{6.650 / 7} &\textbf{5.830 / 7} &\textbf{5.983 / 7} &\textbf{6.440 / 7} &\textbf{6.560 / 7} &\textbf{6.913 / 7} &\textbf{38.377} \\
\bottomrule[1.5pt]
\end{tabular}
}
\caption{The performance of the Qwen-2.5-72B model on bilingual dataset.}
\label{tab:comparison_qwen_zh}
\end{table*}

\begin{table*}[!htp]\centering
\centering
% \small
\scalebox{0.82}{
\begin{tabular}{cccccccc|c}
\toprule[1.5pt]
\multirow{2}{*}{\textbf{Lang.}} &\multirow{2}{*}{\textbf{Method}} &\multirow{2}{*}{\textbf{Empathy}} &\multicolumn{2}{c}{\textbf{Cognitive Distortions}} &\multirow{2}{*}{\textbf{Strategy}} &\multirow{2}{*}{\textbf{Encouragement}} &\multirow{2}{*}{\textbf{Relevance}}  &\multirow{2}{*}{\textbf{Total Score}}\\\cmidrule{4-5}
&&&\textbf{Identification}&\textbf{Reflection} &&&&\\\midrule

\multirow{3}{*}{EN} &Generation &6.055 / 7 &5.267 / 7 &5.161 / 7 &6.059 / 7 &5.549 / 7 &6.718 / 7 &34.810 \\
&PromptCBT  &6.377 / 7 &5.678 / 7 &5.886 / 7 &5.879 / 7 &6.103 / 7 &6.799 / 7 &36.722 \\
&AutoCBT     &\textbf{6.513 / 7} &\textbf{5.780 / 7} &\textbf{5.996 / 7} &\textbf{5.908 / 7} &\textbf{6.227 / 7} &\textbf{6.909 / 7} &\textbf{37.333} \\
\bottomrule[1.5pt]
\end{tabular}
}
\caption{The performance of the Llama-3.1-70B model on bilingual dataset. For a more detailed analysis refer to~\ref{Over-Protection} section.}
\label{tab:comparison_llama_en}
\end{table*}

\paragraph{Human Evaluation}
To identify deeper and more nuanced cognitive distortions, we developed a human evaluation metric in Appendix~\ref{sec:cds}.
Compared to the metrics used in automatic evaluation in Table~\ref{tab:metrics}, human evaluation emphasizes identifying and challenging cognitive distortions more strongly.
Human evaluations include two experiments. In the \textbf{Simple Overall Evaluation (SOE)}, five psychology professionals reviewed all questions in the bilingual dataset and identified the best answer for each question from AutoCBT and baselines.
In the \textbf{Detailed Sampling Evaluation (DSE)}, six psychology professionals evaluate 60 responses across seven dimensions for 10\% of the bilingual dataset, analyzing the strengths and weaknesses of each approach.

\subsection{Results}
The observed scores for the responses in the Chinese portion of the dataset are presented in Table~\ref{tab:comparison_qwen_zh}.
When comparing the effectiveness of Generation and PromptCBT, it is evident that the introduction of CBT core principles significantly enhances the quality of LLMs' responses. Furthermore, based on these principles, AutoCBT generates higher-quality answers than PromptCBT, as shown by its superior performance in 5 out of 6 evaluation metrics.

According to the English portion of the dataset in Table~\ref{tab:comparison_qwen_zh} and Table~\ref{tab:comparison_llama_en}, we observe that AutoCBT outperforms baselines across six evaluation metrics and the overall score in English ability, using the Llama and Qwen models, consistent with its performance in Chinese.

\begin{figure*}[ht]
\centering
\subfloat[Bilingual language performance in SOE.\label{fig:soe}]{
  \includegraphics[width=0.47\textwidth]{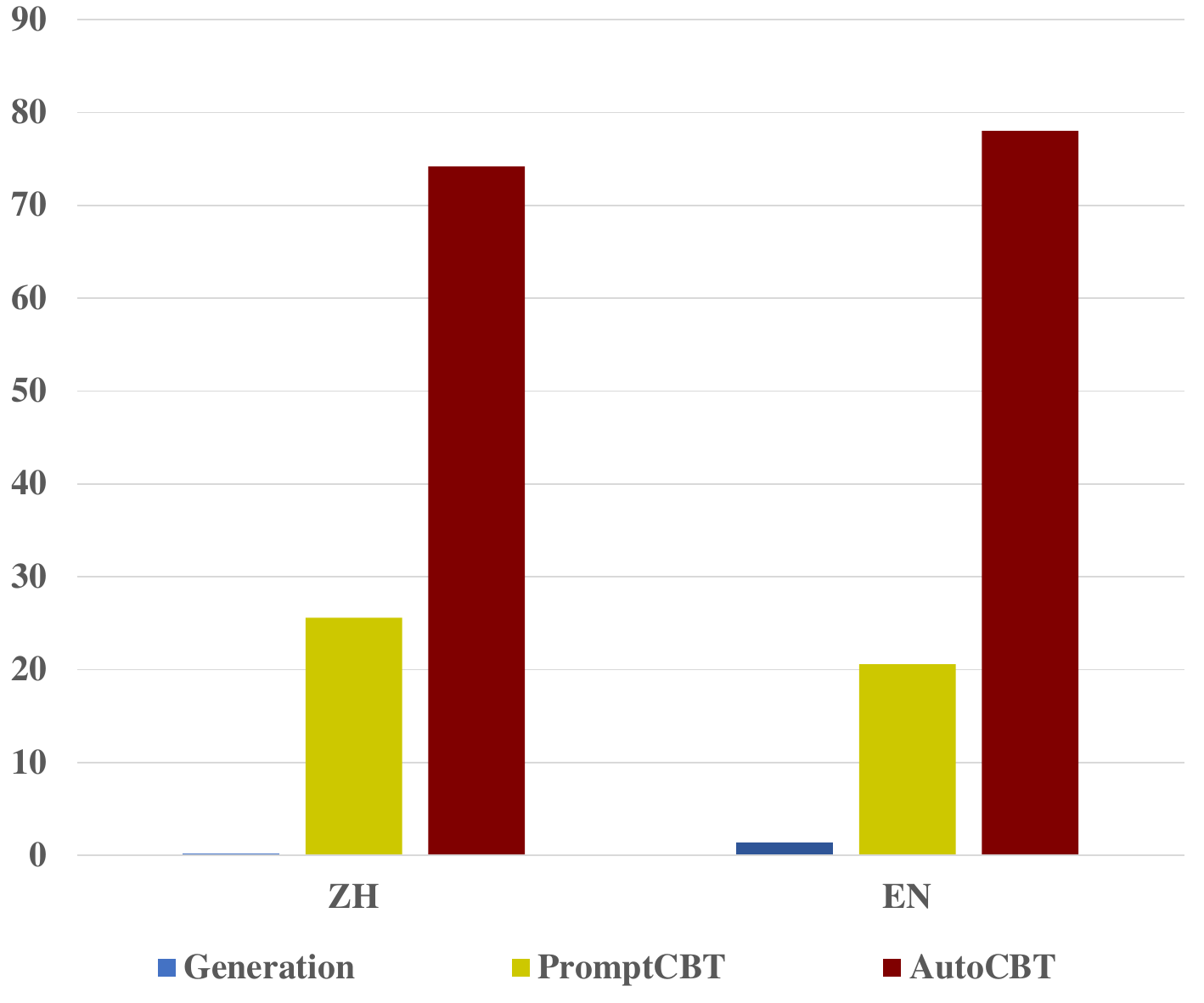}
}\hfil
\subfloat[Bilingual language performance in DSE.\label{fig:dse_zh}]{
  \includegraphics[width=0.47\textwidth]{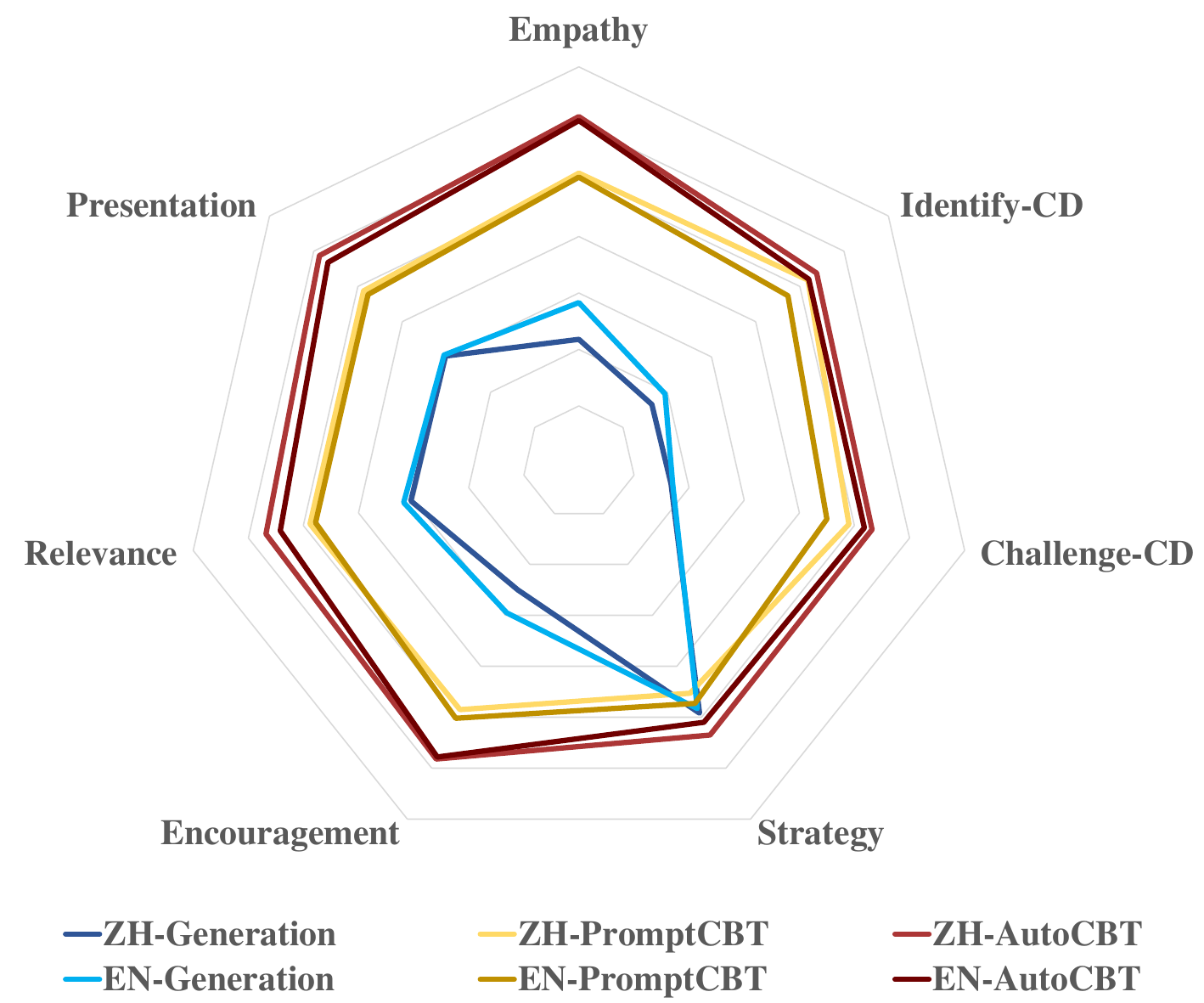}
}\hfil
\caption{AutoCBT generates better answers than both PromptCBT and Generation for over 70\% of the bilingual dataset questions and outperforms both PromptCBT and Generation in identifying and challenging cognitive distortions.}
\label{fig:autocbt_radar}
\end{figure*}

AutoCBT's strong performance is further demonstrated in the human evaluation stage.
Figure~\ref{fig:soe} shows the results of SOE and indicates that AutoCBT provides the best answer for over 70\% of the bilingual dataset questions.
Figure~\ref{fig:dse_zh} shows the results of DSE on identifying and challenging cognitive distortions, psychology professionals prefer AutoCBT’s responses over baseline methods.
AutoCBT outperforms both baseline methods across all seven evaluation dimensions related to cognitive distortions. 

AutoCBT and PromptCBT both use empathetic techniques, but AutoCBT offers warmer support with flexible word choices, which are shaped by cultural differences—respect in Chinese responses and professionalism in English ones.
While both excel in re-description and clarification, AutoCBT’s softer, context-specific tone fosters better emotional validation compared to PromptCBT’s more rigid, academic style, which makes users feel labeled.
Generation suits mild issues, while AutoCBT is ideal for those needing emotional support, showing stronger empathy and encouragement. PromptCBT, though balanced, often lacks clarity, making AutoCBT the best choice for emotional and psychological challenges.

The more specialized psychological perspective on the differences among Generation, PromptCBT, and AutoCBT will be provided in Appendix~\ref{sec:human_analysis}.

\subsection{Discussions} \label{Discussions}
\subsubsection{Validating Generalizability of AutoCBT}
\label{Validating_Generalizability}
To verify the general applicability of AutoCBT and explain how the dynamic routing and supervisory mechanisms in AutoCBT significantly improve the quality of LLM responses, we ensure that the prompt generated by AutoCBT's counsellor agent for the first draft response is consistent with that of PromptCBT.
This approach allows us to observe how these mechanisms enhance the quality of the draft responses.

From the table~\ref{tab:detail_analyze_autocbt_zh_qwen}, we observe that aligning the prompt generated by AutoCBT's counsellor agent for the first draft response with PromptCBT is effective.
In the \textit{Draft and PromptCBT Diff.} column, we observe the difference between the two is minor.
This means that \textbf{AutoCBT's draft response process can simulate the effects of PromptCBT or even other pure prompt method}, as the draft response process of AutoCBT is essentially a customizable prompt.

When the draft response effect of AutoCBT is consistent with PromptCBT, we further compare the score differences between AutoCBT's draft and final responses.
In the \textit{Final and Draft Diff.} column, we observe that the final response scores are significantly higher than the draft response in all metrics.

Therefore, we first demonstrate that AutoCBT's draft response process can achieve the effects of the pure prompt method.
We further demonstrate that a supervisor and dynamic routing mechanism can enhance the quality of psychological counseling Q\&A based on the draft response process, thus validating the effectiveness of the AutoCBT framework for other psychological counseling using pure prompt methods.

\subsubsection{Three Challenges In AutoCBT}
\paragraph{Simultaneous Routing}
In psychological counseling scenarios, the counsellor's decision to engage with the user or supervisor is mutually exclusive, and they cannot choose to "improve dialogue" and "end dialogue" simultaneously. Despite LLMs with 70B+ parameters, their semantic understanding and logic processing remain limited, leading to conflicting routing objectives. Even after emphasizing the distinction between choices, the system may enter a continuous dynamic routing loop where both the user and supervisor repeatedly appear without resolving exclusivity. This persistent issue highlights the inherent limitations of LLMs, making a complete solution challenging.

Therefore, we adopted a different approach: once both the user and supervisor appear as routing targets, indicating that the counsellor is attempting to end the session and exit the loop.

\begin{table*}[!htp]\centering
\scalebox{0.77}{
\begin{tabular}{ccc|cc}
\toprule[1.5pt]
\multirow{2}{*}{\textbf{Metric}}  &\multicolumn{2}{c}{\textbf{Chinese}} &\multicolumn{2}{c}{\textbf{English}}\\\cmidrule{2-5}
&\textbf{Draft and PromptCBT Diff.} &\textbf{Final and Draft Diff.}&\textbf{Draft and PromptCBT Diff.} &\textbf{Final and Draft Diff.}\\\midrule
Empathy &- 0.030 &0.217 &- 0.019 &0.117\\
Identification &0.077 &0.227 &0.052 &0.154\\
Reflection &- 0.054 &0.110 &0.091 &0.201\\
Strategy &0.227 &0.353 &0.040 &0.069\\
Encouragement &- 0.093 &0.224 &0.055 &0.179\\
Relevance &0.050 &0.047 &- 0.033 &0.077\\\midrule
Overall &0.177 &1.177 &0.187 &0.798\\
\bottomrule[1.5pt]
\end{tabular}}
\caption{\textbf{Draft and PromptCBT Diff.} represents the score difference between AutoCBT's draft responses and similar PromptCBT responses; \textbf{Final and Draft Diff.} indicates the improvement in quality score of AutoCBT's final responses compared to AutoCBT's draft responses.}
\label{tab:detail_analyze_autocbt_zh_qwen}
\end{table*}

\begin{table*}[!htp]\centering
\scalebox{0.77}{
\begin{tabular}{cccc|cc}
\toprule[1.5pt]
\multirow{2}{*}{\textbf{Model}}  &\multirow{2}{*}{\textbf{Method}} &\multicolumn{2}{c}{\textbf{Chinese}} &\multicolumn{2}{c}{\textbf{English}} \\\cmidrule{3-6}
&&\textbf{Refused-Questions} &\textbf{Distinct-Refused-Questions} &\textbf{Refused-Questions} &\textbf{Distinct-Refused-Questions}\\\midrule
\multirow{3}{*}{Qwen} 
&Generation  &0 &\multirow{3}{*}{0} &0 &\multirow{3}{*}{0}\\
&PromptCBT   &0 & &0 & \\
&AutoCBT     &0 & &0 &\\\midrule
\multirow{3}{*}{Llama}
&Generation   &/ &\multirow{3}{*}{/} &3 &\multirow{3}{*}{Union(3, 3, 8) = 9}\\
&PromptCBT    &/ & &3 &\\
&AutoCBT      &/ & &$ 8 \rightarrow 2 $ &\\
\bottomrule[1.5pt]
\end{tabular}}
\caption{Rejections by Qwen-2.5-72B and Llama-3.1-70B were analyzed. Llama initially rejected 8 questions, reduced to 2 after AutoCBT's enhancements. In total, Llama rejected 9 unique questions.}
\label{tab:detail_analyze_autocbt_en_llama_refuse_response}
\end{table*}

\paragraph{Confusing Role}
When the user asked the counsellor a question, the counsellor drafted a response but believed it needed improvement and sought advice from supervisors.
However, the supervisor mistakenly generated a response instead of advice, confusing the counsellor, who had expected guidance, not a response.
We speculate that this also stems from the insufficient semantic understanding and instruction-following capabilities of LLMs, especially when the current supervisor finds that the conversation history already contains advice from other supervisor, making role confusion more likely.
We have not completely eliminated the role confusion issue in LLM role-playing scenarios.

Therefore, we modified the prompt to instruct the supervisor agent to start with "Hello counsellor," clearly establishing its role and ensuring it generates responses as a supervisor, not a counsellor.

\paragraph{Routing Loop}
In our agent topology, counsellor send draft responses to supervisors, who provide advice directly without communicating with other supervisors.
In real-world counseling, repeated advice requests on the same issue are uncommon. However, in this system, the counsellor may repeatedly direct requests to the same supervisor due to LLMs' limited semantic understanding and instruction-following capabilities, making it challenging to dynamically avoid redundant selections.

Therefore, we adopted the following approach:
After agent A sends the message to agent B, we remove the directed edge $A \rightarrow B$ in the topology diagram, ensuring that each supervisor is accessed only once by the counsellor.
Assuming there are \(N\) supervisor agents and one user agent, the counsellor can perform up to \(N + 1\) routing operations.

\subsubsection{Over-Protection of Llama}
\label{Over-Protection}
During a psychological counseling simulation, the Llama model refuses to answer nine questions from the English portion of the dataset related to minors, sex, and suicide, a behavior also observed in both AutoCBT and baselines. In contrast, the Qwen model answers all questions of the bilingual dataset. After excluding the nine questions that Llama refuses to answer, we analyze AutoCBT's effectiveness using the Llama model on the remaining 91 questions in Table~\ref{tab:comparison_llama_en} and Table~\ref{tab:detail_analyze_autocbt_en_llama_refuse_response}. Compared with  English results of Table~\ref{tab:comparison_qwen_zh}, Qwen performs significantly better. In baseline scenarios, Llama consistently refuses to answer sensitive questions, fails to recognize the user’s need for psychological support, and significantly harms the user experience.

When deploying online psychological counseling services, Llama's excessive caution can cause users seeking help to feel misunderstood, potentially worsening cognitive distortions.

\section{Conclusion}
In our research, we propose AutoCBT, a general multi-agent framework for CBT-based psychological counseling. This framework introduces dynamic routing and supervisory mechanisms into traditional LLM-based psychological counseling scenarios, significantly enhancing dialogue quality in CBT-based psychological counseling and providing assistance in identifying and challenging users' cognitive distortions. 

We also demonstrate that AutoCBT can significantly improve dialogue quality compared to other purely prompt-based counseling frameworks and explain the mechanism through which AutoCBT achieves this improvement.

During the experiment, we identified deficiencies in LLMs' abilities to follow instructions and comprehend underlying semantics. Subsequently, we analyzed these challenges and proposed solutions to address them.

\section*{Acknowledgements}
We would like to sincerely thank several graduate students—Ruiqi Huang, Yafang Shi, Guiting Chen, Shuangbei Wu, and Yuqing Cui—from the School of Government at Shenzhen University. Their invaluable help with manual annotation and psychological analysis has been essential to our work.

\bibliography{tacl2021, anthology}
\bibliographystyle{acl_natbib}

\onecolumn
\appendix

\section{Evaluation Metrics of Detailed Sampling Evaluation}\label{sec:cds}

The Detailed Sampling Evaluation metrics update the previous automatic evaluation metrics' \textit{Identification} and \textit{Reflection} with \textit{Identify-CD} and \textit{Challenge-CD}. Additionally, the \textit{Presentation} metric is introduced to evaluate the overall performance of the response of the counsellor.

Thus, we retain four metrics identical to those used in the automatic evaluation experiment and introduced three new ones in the Detailed Sampling Evaluation.

\subsection{Consistent Metrics}

\begin{table*}[ht]\centering
\small
\scalebox{1}{
\begin{tabular}{p{3.5cm}p{3.5cm}p{3.5cm}p{3.5cm}}
\toprule[1.5pt]
\multicolumn{4}{l}{\textbf{Consistent with previous metrics}} \\\midrule
Empathy &Strategy &Encouragement &Relevance\\
\bottomrule[1.5pt]
\end{tabular}
}
\caption{The four metrics are consistent with the previous metrics, which were automatically evaluated.}
\label{tab:metrics_cd}
\end{table*}

\subsection{New Metrics}

\begin{table*}[ht]\centering
\small
\scalebox{1}{
\begin{tabular}{lp{4.4cm}p{6.5cm}c}
\toprule[1.5pt]
\textbf{Perspective} &\textbf{Description} &\textbf{Criterion} &\textbf{Score} \\\midrule
\multirow{6}{*}{\color{black}{Identify-CD}} &\multirow{6}{4.5cm}{Identify potential cognitive distortions of the user through the description of the problem in the dialogue} & 2.1 Has the cognitive distortion phenomenon of users been identified? &\multirow{6}{*}{7} \\ 
~ & ~ & 2.2 Does it help users recognize distorted beliefs? & ~ \\
~ & ~ & 2.3 Has cognitive distortion been explained from a psychological perspective? & ~ \\ \midrule
\multirow{9}{*}{\color{black}{Challenge-CD}} &\multirow{9}{4.5cm}{Ask open-ended questions to encourage the user to reconsider or reflect on their initial thoughts or beliefs} & 3.1 Does it help users think and challenge these distorted beliefs? &\multirow{9}{*}{7} \\ 
~ & ~ & 3.2 Have you raised open-ended questions that are helpful for deeper thinking? & ~ \\
~ & ~ & 3.3 Has psychological counseling technology been integrated? & ~ \\
~ & ~ & 3.4 Does the guided reflection correspond to the cognitive distortions that visitors may have? \\\midrule
\multirow{5}{*}{\color{black}{Presentation}} &\multirow{5}{4.5cm}{Evaluate the overall performance of the response of counsellor} & 7.1 Is the overall language style close to the image of counsellor? &\multirow{5}{*}{7} \\ 
~ & ~ & 7.2 Is the information expressed clearly? & ~ \\
~ & ~ & 7.3 Have you flexibly applied some psychological counseling techniques? & ~ \\
\bottomrule[1.5pt]
\end{tabular}
}
\caption{The three metrics are inconsistent with the previous metrics, which were automatically evaluated.}
\label{tab:metrics_cd}
\end{table*}

\section{Human Analysis of Detailed Sampling Evaluation}\label{sec:human_analysis}

In the Detailed Sampling Evaluation, six psychological experts meticulously analyze the differences between AutoCBT and baselines, guided by the principles of cognitive distortions, and summarize the following analytical content:

\begin{table*}[ht]\centering
% \small
\scalebox{0.8}{
\begin{tabular}{lp{14.5cm}}
\toprule[1.5pt]
\textbf{Perspective} &\textbf{Human Analysis} \\\midrule
\multirow{19}{*}{Empathy and Encouragement} &{The responses from AutoCBT are logically similar to those from PromptCBT, both begin with empathetic techniques to convey understanding and validate the user's challenges before moving to structured and logical assessments and recommendations.
Both approaches generally demonstrate an accurate understanding of the user's concerns and challenges.
However, AutoCBT provides slightly more emotional support, creating an overall warmer impression.
Its responses integrate empathetic techniques more smoothly and maintain a consistent empathetic tone.
Additionally, two specific aspects were observed.
First, AutoCBT demonstrates more flexibility in word choice compared to PromptCBT.
This marks a significant improvement over the formulaic responses typically associated with previous LLMs.
Furthermore, likely due to cultural differences in counseling model training, AutoCBT's approach to creating a “safe environment for the user” varies between its Chinese and English responses.
In the Chinese context, it emphasizes respect, attentiveness, and ensures the user feels valued, respected, and heard.
In English, however, it emphasizes professionalism with phrases like, “I’ll view your issue from a non-judgmental perspective,” aligning with the clear boundaries often emphasized in Western society.
In Chinese practice, these boundaries are generally less pronounced to avoid creating user apprehension.} \\\midrule
\multirow{25}{*}{Cognitive Distortion} &{Both AutoCBT and PromptCBT effectively identify and analyze users' cognitive distortions; however, Generation’s responses contain minimal content on this aspect.
There is a notable gap between AutoCBT and PromptCBT in further challenging cognitive distortions, primarily in their integration with the client’s context.
PromptCBT’s guided reflection can feel rigid, and some responses may make users feel interrogated.
In contrast, AutoCBT’s recognition and reflection are well-aligned with users’ specific contexts, using softer, gentler language that guides users to examine the rationality of their core beliefs from different perspectives.

Both AutoCBT and PromptCBT responses exhibit re-description, summarization, and conceptual clarification of user questions, with AutoCBT applying these techniques more extensively.
We see this as a key advantage of LLM-based psychological counseling responses.
Re-description not only demonstrates that the "Counselor Agent" genuinely understands the user’s issue but also enhances the credibility of “I can understand you,” helping users feel their emotions are acknowledged.
Additionally, users experiencing psychological and emotional challenges often have confused thoughts.
Techniques like re-description, summarization, and clarification assist users in clarifying their logical thinking and focusing on the issues they seek to resolve.
Additionally, in vocabulary explanation, AutoCBT uses a more approachable and conversational language style, while PromptCBT tends toward academic expressions.
PromptCBT often uses more specialized psychological terms, which can inadvertently make users feel “labeled” and lead to self-criticism.
For instance, PromptCBT might use terms like “catastrophizing thinking,” potentially leading users to think, “I’m really bad.”
Similar issues occasionally appear in AutoCBT’s responses but with less frequency than in PromptCBT’s.
In real-life counseling, practitioners carefully use professional terminology, especially with clients experiencing significant psychological challenges.
They often use more tactful language when conveying serious-sounding terms, a strength in which AutoCBT excels.} \\\midrule
\multirow{19}{*}{Usefulness of the strategy} &{Based on its performance, we believe Generation is suitable primarily for users with mild emotional issues and a clear objective of finding problem-solving methods. However, its mechanical and rigid language is less appropriate for users needing psychological and emotional support.
For users experiencing emotional confusion or in a suboptimal or unhealthy psychological state, AutoCBT is recommended. AutoCBT's performance more closely resembles that of a psychological counselor, providing greater empathy and respect in its language.
PromptCBT’s positioning lies between the other two; it employs more academic language that may seem diagnostic rather than consultative, lacking clear explanations for users.
Generation offers the widest range of strategies among the three, providing users with diverse choices. However, its strategies are often vague, with broad, generic explanations that lack specific responses to users' challenges, leading to lower overall relevance.
AutoCBT and PromptCBT incorporate user-specific contexts to better address their needs. Of the two, AutoCBT performs better, showing stronger empathy and encouragement in its language, and creating a more genuine dialogue with users.
When proposing potentially sensitive strategies, like suggesting users seek professional counseling, AutoCBT uses caring language paired with empathy and encouragement, reducing visitors' resistance.
In some responses, AutoCBT anticipates potential obstacles in implementing strategies and provides timely encouragement, offering empathetic support for users with psychological or emotional challenges.} \\
\bottomrule[1.5pt]
\end{tabular}
}
\caption{Human analysis of the DSE.}
\label{tab:human_analysis}
\end{table*}

\iftaclpubformat

\onecolumn

\fi
\end{CJK*}
\end{document}